\documentclass[11pt,a4paper]{article}
\usepackage[hyperref]{naaclhlt2019}
\usepackage{times}
\usepackage{latexsym}
\usepackage{color}
\usepackage{dsfont}
\usepackage{amsmath}
\usepackage{amssymb}
\usepackage{graphicx}
\usepackage{url}
\usepackage{booktabs}

\aclfinalcopy

\title{Simple Attention-Based Representation Learning for \\ Ranking Short Social Media Posts}

\author{Peng Shi,$^1$ Jinfeng Rao,$^2$\thanks{\hspace{0.1cm}~Work done at the University of Maryland, College Park.}~~\and Jimmy Lin$^1$\vspace{0.1cm}\\
$^1$ David R. Cheriton School of Computer Science, University of Waterloo\\
$^2$ Facebook AI\vspace{0.1cm}\\
{\tt \{peng.shi, jimmylin\}@uwaterloo.ca}, {\tt raojinfeng@fb.com}}

\date{}

\begin{document}
\maketitle
\begin{abstract}
 This paper explores the problem of ranking short social media posts
with respect to user queries using neural networks. Instead of
starting with a complex architecture, we proceed from the bottom up
and examine the effectiveness of a simple, word-level Siamese
architecture augmented with attention-based mechanisms for capturing
semantic ``soft'' matches between query and post tokens. 
Extensive experiments on datasets from the TREC Microblog Tracks
show that our simple models not only achieve better effectiveness 
than existing approaches that are far more complex or exploit a
more diverse set of relevance signals, but are also much faster.
Implementations of our \textbf{samCNN}~(\textbf{S}imple \textbf{A}ttention-based \textbf{M}atching CNN) models are shared with the community to support future work.\footnote{\url{https://github.com/Impavidity/samCNN}}
\end{abstract}

\section{Introduction}

Despite a large body of work on neural ranking models for
``traditional'' {\it ad hoc} retrieval over web pages and newswire
documents~\citep{huang2013learning,shen2014learning,guo2016deep,xiong2017end,mitra2017learning,pang2017deeprank,dai2018convolutional,mcdonald2018deep},
there has been surprisingly little work~\cite{rao2017integrating} on applying neural networks to
searching short social media posts such as tweets on
Twitter. \citet{rao2018tweet} identified short document length,
informality of language, and heterogeneous relevance signals as main
challenges in relevance modeling, and proposed the first neural model specifically
designed to handle these characteristics. Evaluation on a number of
datasets from the TREC Microblog Tracks demonstrates state-of-the-art
effectiveness as well as the necessity of different model components
to capture a multitude of relevance signals.

\begin{figure}[t]
    \centering
    \includegraphics[width=0.35\paperwidth]{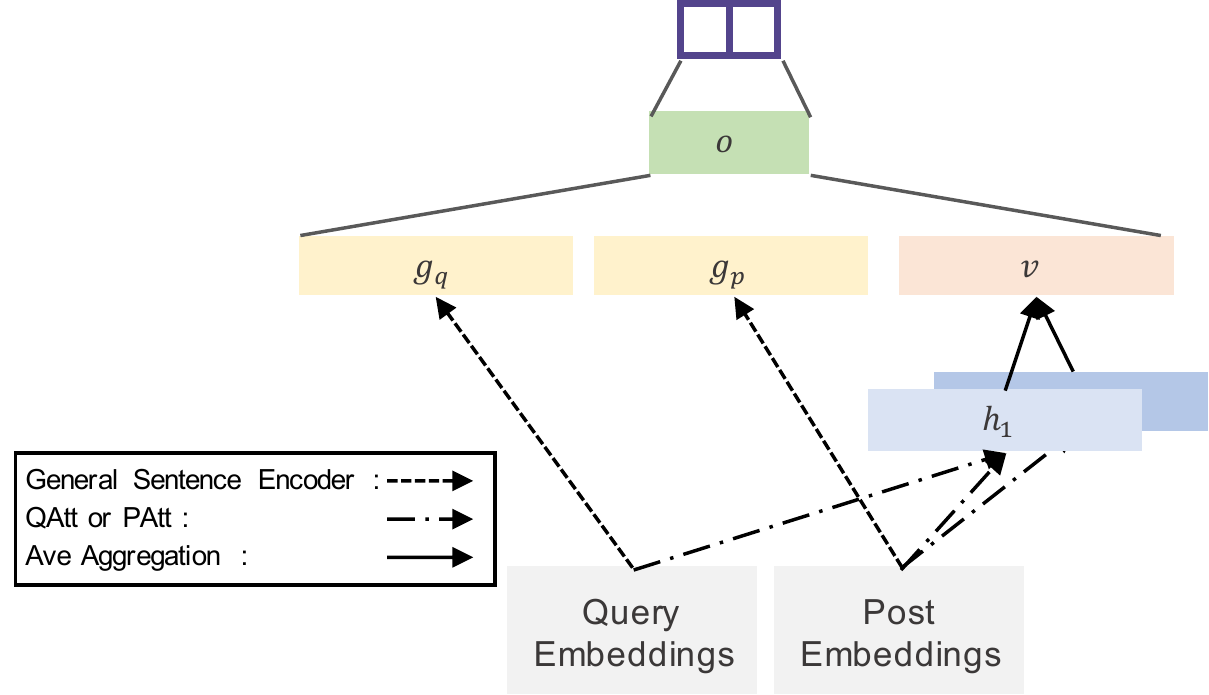}
    \caption{Our model architecture:\ a general sentence encoder is
      applied on query and post embeddings to generate $\textbf{g}_q$
      and $\textbf{g}_p$; an attention encoder is applied on post
      embeddings to generate variable-length query-aware features
      ${\textbf{h}_i}$.  These features are further aggregated to
      yield $\textbf{v}$, which feeds into the final prediction.}
    \label{SystemFigure}
\end{figure}

\begin{figure*}[t]
	\centering
    \includegraphics[width=0.85\textwidth]{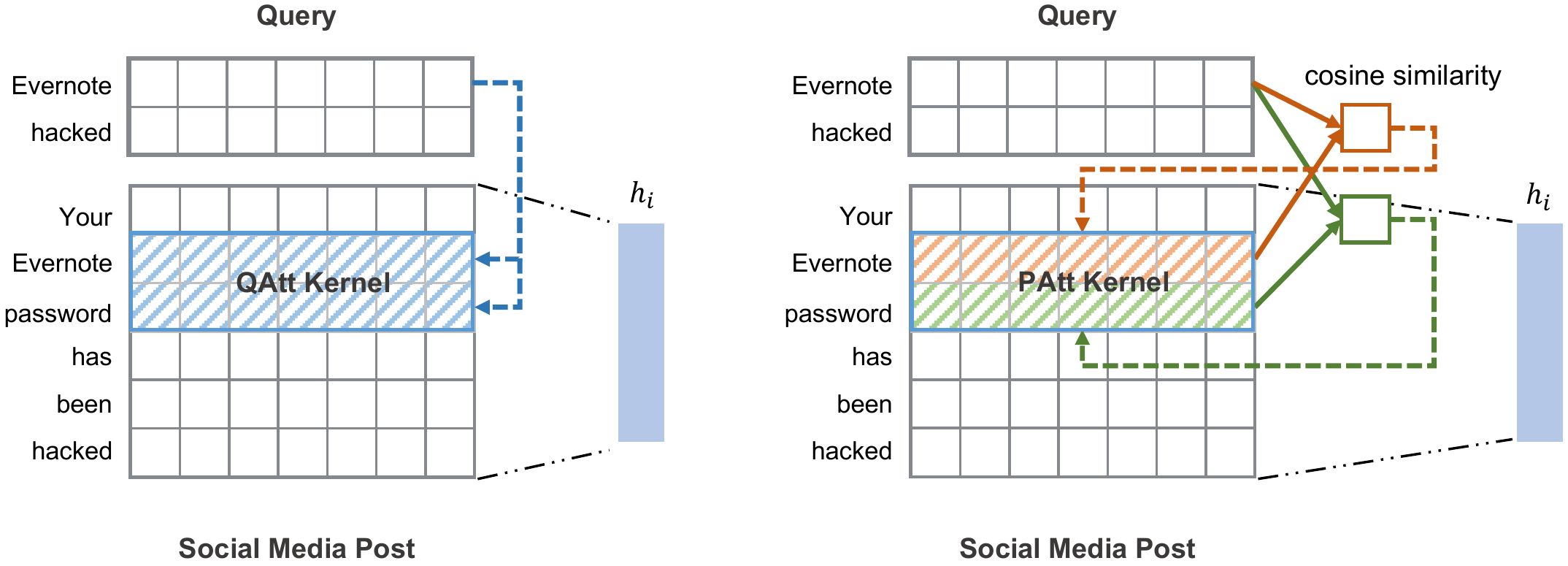}
    \caption{The Query-Aware Attention (QAtt) architecture on the left and the Position-Aware Attention (PAtt) architecture on the right.
In both, we construct $F$ convolutional kernels for {\it each} query token (here, one kernel for the query token `Evernote' is visualized).
In QAtt, the query token embedding is directly ``injected'' into the kernel via element-wise product (blue dotted arrows).
In PAtt, cosine similarity between the query token and tokens in the post within the convolution window are used as attention weights in the kernel.}
    \label{kernelFigure}
\end{figure*}

In this paper, we also examine the problem of modeling relevance for ranking short social media posts, but from a complementary perspective. 
As \citet{weissenborn2017making} notes, most systems are built in a \textit{top-down} process:\ authors propose a complex architecture and then validate design decisions with ablation experiments.
However, such experiments often lack comparisons to strong baselines, which raises the question as to whether model complexity is empirically justified.
As an alternative, they advocate a \textit{bottom-up} approach where architectural complexity is gradually increased.
We adopt exactly such an approach, focused exclusively on word-level modeling.
As shown in Figure~\ref{SystemFigure}, we examine variants of a simple, generic architecture that has emerged as ``best practices'' in the NLP community for tackling modeling problems on two input sequences:\
a Siamese CNN architecture for learning representations over both inputs (a query and a social media post in our case), followed by fully-connected layers that produce a final relevance prediction~\citep{Severyn_Moschitti_SIGIR2015,he2016umd,rao2016noise}, which we refer to as a General Sentence Encoder in Section~\ref{sec:representation-learning}. 
Further adopting best practices, we incorporate query-aware convolutions with an average aggregation layer in the representation learning process. 

Recently, a number of researchers~\citep{conneau2017supervised,mohammed2017strong} have started to reexamine simple baselines and found them to be highly competitive with the state of the art, especially with proper tuning.
For example, the InferSent approach~\cite{conneau2017supervised} uses a simple BiLSTM with max pooling that achieves quite impressive accuracy on several classification benchmarks. 
Our contribution is along similar lines, where we explore simple yet highly effective models for ranking social media posts, to gain insights into query--post relevance matching using standard neural architectures.  
Experiments with TREC Microblog datasets show that our best model not only achieves better effectiveness than existing approaches that leverage more signals, but also demonstrates 4$\times$ speedup in model training and inference compared to a recently-proposed neural model.

\section{Model}

Our model comprises a representation learning layer with convolutional encoders and another simple aggregation layer.
These architectural components are described in detail below.

\subsection{Representation Learning Layer}
\label{sec:representation-learning}

\noindent \textbf{General Sentence Encoder:}\
The general sentence encoder uses a standard convolutional layer with randomly initialized kernels to learn
semantic representations for text. More formally, given query $q$ and post $p$ as sentence inputs, 
we first convert them to embedding matrices $\textbf{Q}$ and $\textbf{P}$ through an embedding lookup layer, where $\textbf{Q} \in \mathbb{R}^{n \times d}$ and $\textbf{P} \in \mathbb{R}^{m \times d}$, 
$d$ is the dimension of embeddings, and $n$ and $m$ are the number of tokens in $q$ and $p$, respectively.
Then we apply a standard convolution operation with kernel window size $k$ over the embedding matrix
$\textbf{Q}$ and $\textbf{P}$. The convolution operation is parameterized by 
a weight term $\textbf{W} \in \mathbb{R}^{F \times k \times d}$ and a bias term $\textbf{$b_w$} \in \mathbb{R}^{F}$, where $F$ is the number of convolutional kernels. This generates semantic representation $\textbf{O}_q \in \mathbb{R}^{n \times F}$ and $\textbf{O}_p \in \mathbb{R}^{m \times F}$, on which max pooling and an MLP are applied to
obtain query representation $\textbf{g}_q \in \mathbb{R}^d$ and post representation $\textbf{g}_p \in \mathbb{R}^d$.

The weakness of the kernels in the general sentence encoder is that they do not incorporate knowledge from the query when attempting to capture feature patterns from the post.
Inspired by attention mechanisms~\cite{bahdanau2014neural}, we propose two novel approaches to incorporate query information when encoding the post representation, which we introduce below.

\smallskip \noindent \textbf{Query-Aware Attention Encoder (QAtt):}\
In QAtt (Figure~\ref{kernelFigure}, left), for each query token, we construct a token-specific convolutional kernel to ``inject'' the query information.
Unlike methods that apply attention mechanisms after the sentence representations are generated~\cite{bahdanau2014neural, seo2016bidirectional}, our approach aims to model the representation learning process jointly with an attention mechanism.

Formally, for each query token $t_q$, the QAtt kernel $\mathbf{W}_{\textrm{QAtt}}^{t_q}$ is composed as follows:
\begin{equation}
    \mathbf{W}_{\textrm{QAtt}}^{t_q} = \mathbf{U} \otimes \mathbf{Q}_{t_q}
\end{equation}
\noindent where $\mathbf{U} \in \mathbb{R}^{F \times k \times d }$ represents trainable parameters, $\mathbf{Q}_{t_q}$ is the embedding of token $t_q$ with 
size $\mathbb{R}^d$ and $\mathbf{W}_{\textrm{QAtt}}^{t_q} \in \mathbb{R}^{F \times k \times d }$. The element-wise product $\otimes$ is applied between the token embedding
$\mathbf{Q}_{t_q}$ and the last dimension of kernel weights $\mathbf{U}$.
In other words, we create $F$ convolutional kernels for {\it each} query token, where each kernel is ``injected'' with the embedding of that query token via element-wise product.
Figure~\ref{kernelFigure} (left) illustrates one kernel for the query token `Evernote', where element-wise product is represented by blue dotted arrows.
When a QAtt token-specific kernel is applied, a window slides across the post embeddings $\textbf{P}$ and learns soft matches to each query token to generate query-aware representations.

On top of the QAtt kernels, we apply max-pooling and an MLP to produce a set of post representations $\{\textbf{h}_i\}$, with each $\textbf{h}_i  \in \mathbb{R}^d$ standing for the representation learned from query token $t_{q_i}$.

\smallskip \noindent \textbf{Position-Aware Attention Encoder (PAtt):}\
In the QAtt encoder, token-specific kernels learn soft matches to the query.
However, they still ignore positional information when encoding the post semantics, which has been shown to be effective for sequence modeling~\cite{gehring2017convolutional}.
To overcome this limitation, we propose an alternative attention encoder that captures positional information through interactions between query embeddings and post embeddings.

Given a query token $t_q$ and the $j$-th position in post $p$,
we compute the interaction scores by taking the cosine similarity between the word
embeddings of token $t_q$ and post tokens $t_{p_{j: j+k-1}}$ from position $j$ to $j+k-1$:
\begin{equation} 
S_j = [\textrm{cos}(t_q, t_{p_j}); ...; \textrm{cos}(t_q, t_{p_{j+k-1}})]
\end{equation}
\noindent where $S_j \in \mathbb{R}^{k \times 1}$ and $k$ is the width of the convolutional kernel we are learning.
That is, for each token in the post within the window, we compute its cosine similarity with query token $t_q$.
We then convert the similarity vector $S_j$ into a matrix:
\begin{equation}
\hat{S}_j = S_j \cdot \mathds{1}, \hat{S}_j \in \mathbb{R}^{k \times d}
\end{equation}
\noindent where $\mathds{1} \in \mathbb{R}^{1 \times d}$ with each element set to 1. Finally, the PAtt
convolutional kernel for query token $t_q$ at the $j$-th position is constructed as:
\begin{equation}
 \mathbf{W}_{\textrm{PAtt}}^{t_q, j} = \mathbf{V} \otimes \hat{S}_j
\end{equation}
\noindent where $\mathbf{V} \in \mathbb{R}^{F \times k \times d }$ represents the trainable parameters. The element-wise product $\otimes$ is applied between the attention weights
$\hat{S}_j$ and the last two dimensions of kernel weights $\mathbf{V}$.

Conceptually, this operation can be thought as adding a soft attention weight (with values in the range of $[0, 1]$) to each convolutional kernel, where the weight is determined by the cosine similarity between the token from the post and a particular query token; since cosine similarity is a scalar, we fill in the value in all $d$ dimensions of the kernel, where $d$ is the size of the word embedding.
This is illustrated in Figure~\ref{kernelFigure} (right), where we show one kernel of width two for the query token `Evernote'. The brown (green) arrows capture cosine similarity between the query token `Evernote' and the first (second) token from the post in the window. These values then serve as weights in the kernels, shown as the hatched areas.
Similar to QAtt, the PAtt encoder with max-pooling and an MLP generates a set of post representations $\{\textbf{h}_i\}$, with each $\textbf{h}_i$ standing for the representation learned from query token $t_{q_i}$.

It is worth noting that both the QAtt and PAtt encoders have no extra parameters over a general sentence encoder. 
However, incorporating the query-aware and position-aware information 
enables more effective representation learning, as our experiments show later.
The QAtt and PAtt encoders can also be used as plug-in modules in any standard convolutional architecture to learn query-biased representations.

\subsection{Aggregation Layer}
\label{sec:aggregation}

After the representation layer, a set of vectors $\{\textbf{g}_q, \textbf{g}_p,\{\textbf{h}_i\}\}$ is obtained.
Because our model yields different numbers of $\textbf{h}_i$ with queries of different lengths, further aggregation is needed to output a global feature $\textbf{v}$.
We directly average all vectors $\mathbf{v} = \frac{1}{N_{q}}\sum{\mathbf{h_i}}$ as the aggregated feature, where $N_q$ is the length of the query.

\subsection{Training}

To obtain a final relevance score, the feature vectors $\textbf{g}_q$, $\textbf{g}_p$, and $\textbf{v}$ are concatenated and fed into an MLP with ReLU activation for dimensionality reduction to obtain $\textbf{o}$, followed by batch normalization and fully-connected layer and softmax to output the final prediction.
The model is trained end-to-end with a Stochastic Gradient Decent optimizer using negative log-likelihood loss.

\begin{table}[t]
	\center
	\small
	\begin{tabular}{p{4.2em} rrrr}
		\toprule
		\textbf{Year} & \textbf{2011}  & \textbf{2012}  & \textbf{2013}  & \textbf{2014}  \\ \toprule
		\textbf{\# queries}   & 49    & 60    & 60    & 55    \\ 
		\textbf{\# tweets} & 39,780 & 49,879 & 46,192 & 41,579 \\ 
		\textbf{\# relevant}  & 1,940  & 4,298  & 3,405  & 6,812  \\ 
		\textbf{\%relevant}   & 4.87  & 8.62  & 7.37  & 16.38 \\ 
		\bottomrule
	\end{tabular}
	\caption{Statistics of TREC MB 2011--2014 datasets. }
	\label{microblog:statistics}
\end{table}

\begin{table}[t]
	\center
	\small
	\begin{tabular}{p{6.5em} p{2.5em}|p{7em}p{2em}}
		\toprule
		\textbf{Param} & \textbf{Value}  & \textbf{Param}  & \textbf{Value}   \\ \toprule
		Embedding size  & 300    & $k$ & 0.05      \\ 
		Hidden size  & 200  & Final hidden size & 100  \\ 
		Kernel number & 250  & Dropout ratio  & 0.5   \\ 
		Kernel size & 2 &  Learning rate & 0.03\\
		\bottomrule
	\end{tabular}
	\caption{Hyperparameters for our models. GloVe~\citep{pennington2014glove} embeddings are used and fine-tuned during training. Unknown words are initialized from a uniform distribution $[-k, k]$.}
	\label{hyperparameter}
\end{table}

\section{Experimental Setup}

\begin{table*}[!ht]
	\centering
	\small
	\begin{tabular}{p{0.5em} p{1.2em} p{3.2em} p{3.2em} p{3.2em} p{3.2em} p{3.2em} p{3.2em} p{3.2em} p{3.2em}} 
		\toprule 
		& & \multicolumn{2}{c}{\textbf{2011}} & \multicolumn{2}{c}{\textbf{2012}} & \multicolumn{2}{c}{\textbf{2013}} & \multicolumn{2}{c}{\textbf{2014}} \\
		  &  & \bf{P30} & \bf{AP} & \bf{P30} & \bf{AP} & \bf{P30} & \bf{AP} & \bf{P30} & \bf{AP} \\
                \toprule
       		\multicolumn{2}{l}{\textbf{Our Models}} \\
		1 & \multicolumn{1}{l}{BiCNN} &  0.2129  & 0.1634 & 0.2028 & 0.1176 &  0.2367 & 0.1284 & 0.3788 & 0.2557 \\
		2 & \multicolumn{1}{l}{BiCNN+QAtt}  &  $0.3966^{1}$ & $0.3586^{1}$ & $0.3904^{1}$ & $0.2376^{1}$ & $0.4861^{1}$ & $0.2696^{1}$ & $0.6388^{1}$ & $0.4226^{1}$ \\
		3 & \multicolumn{1}{l}{BiCNN+PAtt}  & $0.4469^{1,2}_{}$ & $0.4135^{1,2}_{5}$ & $0.4017^{1}_{5}$ & $0.2413^{1,5}$ & $0.5167^{1,2}_{5}$ & $0.2817^{1,2}$ & $0.6642^{1,2}$ & $0.4351^{1,2}_{5}$ \\
		4 & \multicolumn{1}{l}{BiCNN+PAtt+QL}  & $\textbf{0.4735}^{1{\text -}3}_{5{\text -}7}$ & $\textbf{0.4346}^{1{\text -}3}_{5,6}$ & $\textbf{0.4164}^{1,2}_{5,6}$ & $\textbf{0.2516}^{1{\text -}3}_{5}$ & $0.5256^{1{\text -}3}_{5,6}$ & $\textbf{0.2965}^{1{\text -}3}_{5}$ & $\textbf{0.6752}^{1,2}$ & $\textbf{0.4522}^{1{\text -}3}_{5,7}$ \\	
		\midrule
		\multicolumn{2}{l}{\textbf{Existing Models}} \\
		5 & \multicolumn{1}{l}{QL} & $0.4000^{1}$ & $0.3576^{1}$ & $0.3311^{1}$ & $0.2091^{1}$ & $0.4450^{1}$ & $0.2532^{1}$ & $0.6182^{1}$ & $0.3924^{1}$ \\
		6 & \multicolumn{1}{l}{RM3} & $0.4211^{1}$ & $0.3824^{1}$ & $0.3452^{1}$ & $0.2342^{1}$ & $0.4733^{1}$ & $0.2766^{1}$ & $0.6339^{1}$ & $0.4480^{1}$ \\
		7 & \multicolumn{1}{l}{MP-HCNN(+URL)} & $0.4075^{1,2}$ & $0.3832^{1,2}$ & $0.3689^{1,5}$ & $0.2337^{1,5}$ & $0.5222^{1,2}_{5}$ & $0.2818^{1,2}_{5}$ & $0.6297^{1}$ & $0.4304^{1}$ \\
		8 & \multicolumn{1}{l}{MP-HCNN(+URL)+QL} & $0.4293^{1,2}_{5}$ & $0.4043^{1,2}_{5,6}$ & $0.3791^{1,5}_{6}$ & $0.2460^{1,5}$ & $\textbf{0.5294}^{1{\text -}3}_{5,6}$ & $0.2896^{1,2}_{5}$ & $0.6394^{1}$ & $0.4420^{1,5}$ \\
		\bottomrule
	\end{tabular}
	\caption{Results of various models on the TREC Microblog Tracks datasets.
Models 5--8 are copied from~\citet{rao2018tweet}; note that MP-HCNN exploits URL information (+URL).
Models with +QL include interpolation with the QL baseline.
BiCNN denotes our general sentence encoder architecture, with either query-aware attention (QAtt) or position-aware attention (PAtt).
Superscripts and subscripts indicate the row indexes for which a metric difference is statistically significant at $p < 0.05$.}
	\label{microblog:results}
\end{table*}

\noindent \textbf{Datasets and Hyperparameters.}
Our models are evaluated on four tweet test collections from the TREC 2011--2014 Microblog (MB) Tracks~\citep{ounis2011overview,soboroff2012overview,lin2013overview,lin2014overview}. 
Each dataset contains around 50--60 queries; detailed statistics are shown in Table~\ref{microblog:statistics}.
As with~\citet{rao2018tweet}, we evaluated our models in a reranking task, where the inputs are up to the top 1000 tweets retrieved from ``bag of words'' ranking using query likelihood (QL).
We ran four-fold cross-validation split by year (i.e., train on three years' data, test on one year's data) and followed \citet{rao2018tweet} for sampling validation sets.
For metrics, we used average precision (AP) and precision at rank 30 (P30).
We conducted Fisher's two-sided, paired randomization tests~\citep{smucker2007comparison} to assess statistical significance at $p < 0.05$.  
The best model hyperparameters are shown in Table~\ref{hyperparameter}. 

\smallskip \noindent \textbf{Baselines.}
On top of QL, RM3~\citep{Abdul-Jaleel04} provides strong non-neural results using pseudo-relevance feedback.
We also compared against MP-HCNN~\citep{rao2018tweet}, the first neural model that captures specific characteristics of social media posts, which improves over many previous neural models, e.g., K-NRM~\citep{xiong2017end} and DUET~\citep{mitra2017learning}, by a significant margin.
To the best of our knowledge, \citet{rao2018tweet} is the most effective neural model to date.
We compared against two variants of MP-HCNN; MP-HCNN+QL includes a linear interpolation with QL scores.

\section{Results and Discussion}

Table~\ref{microblog:results} shows the effectiveness of all variants of our model, compared against previous results copied from~\citet{rao2018tweet}.
Model~1 illustrates the effectiveness of the basic BiCNN model with a kernel window size of two; combining different window sizes~\cite{kim2014convolutional} doesn't yield any improvements.
It appears that this model performs worse than the QL baseline.

Comparing Model~2 to Model~1, we find that query-aware kernels contribute significant improvements, achieving effectiveness comparable to the QL baseline.
With Model~3, which captures positional information with the position-aware encoder, we obtain competitive effectiveness compared to Model~8, the full MP-HCNN model that includes interpolation with QL.
Note that Model 8 leverages additional signals, including URL information, character-level encodings, and external term features such as tf--idf. 
With Model~4, which interpolates the position-aware encoder with QL, we obtain state-of-the-art effectiveness.

\begin{figure}[t]
	\centering
	\includegraphics[width=0.45\textwidth]{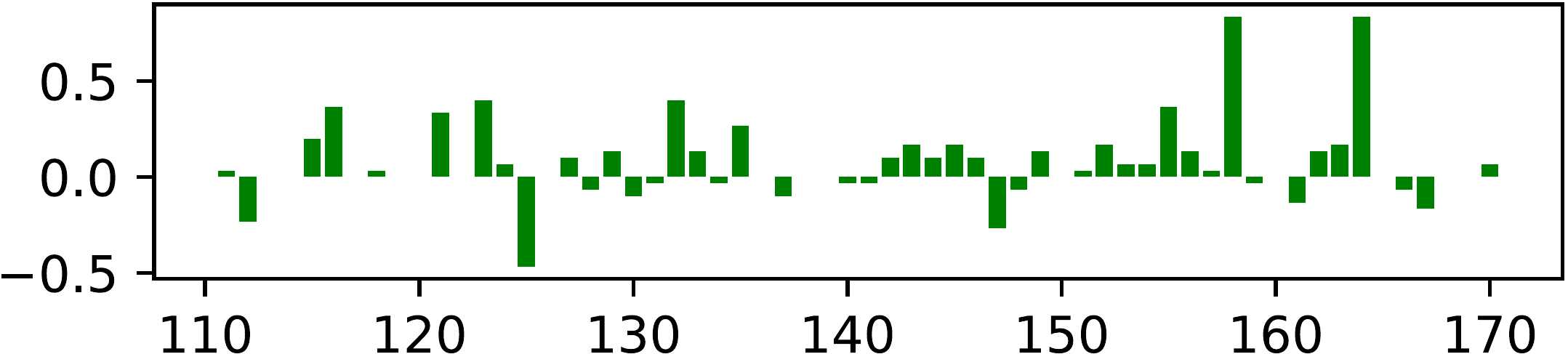}
	\caption{Per-query AP differences between PAtt and QL on TREC 2013 (queries 111--170).}
	\label{per-topic-2013}
\end{figure}

\begin{figure*}[t]
	\centering
	\includegraphics[width=0.32\textwidth]{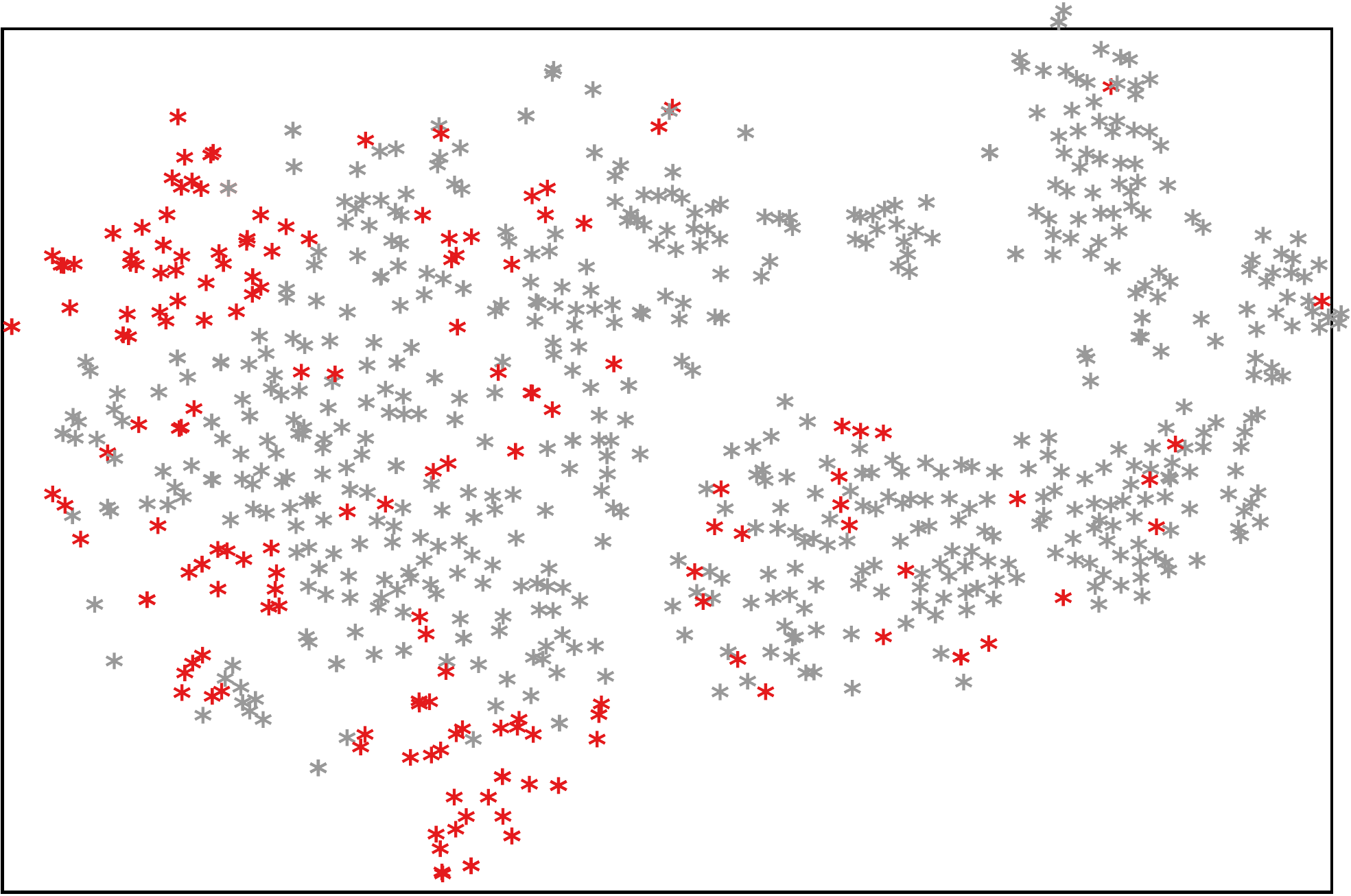}
	\includegraphics[width=0.32\textwidth]{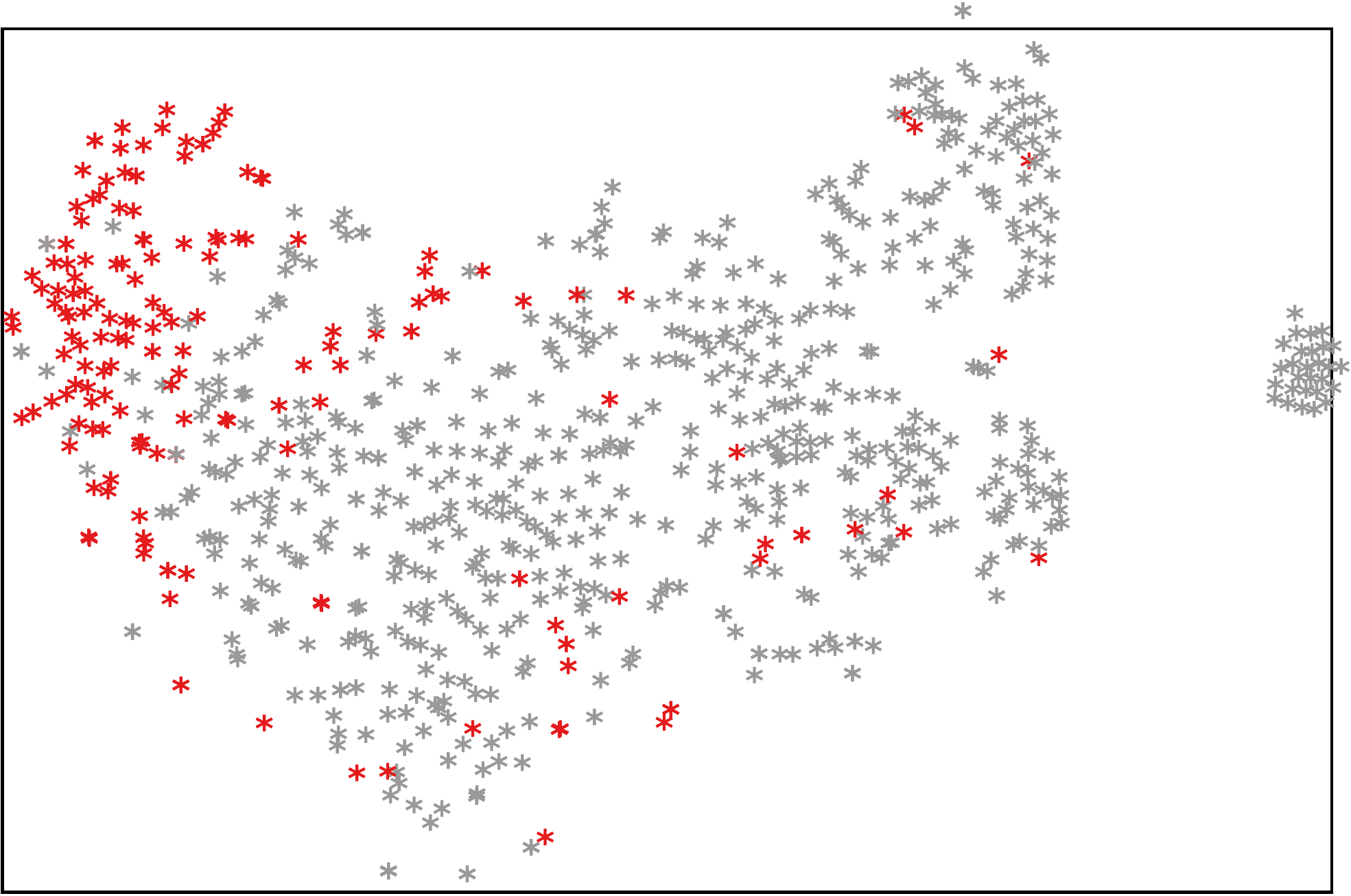}
	\includegraphics[width=0.32\textwidth]{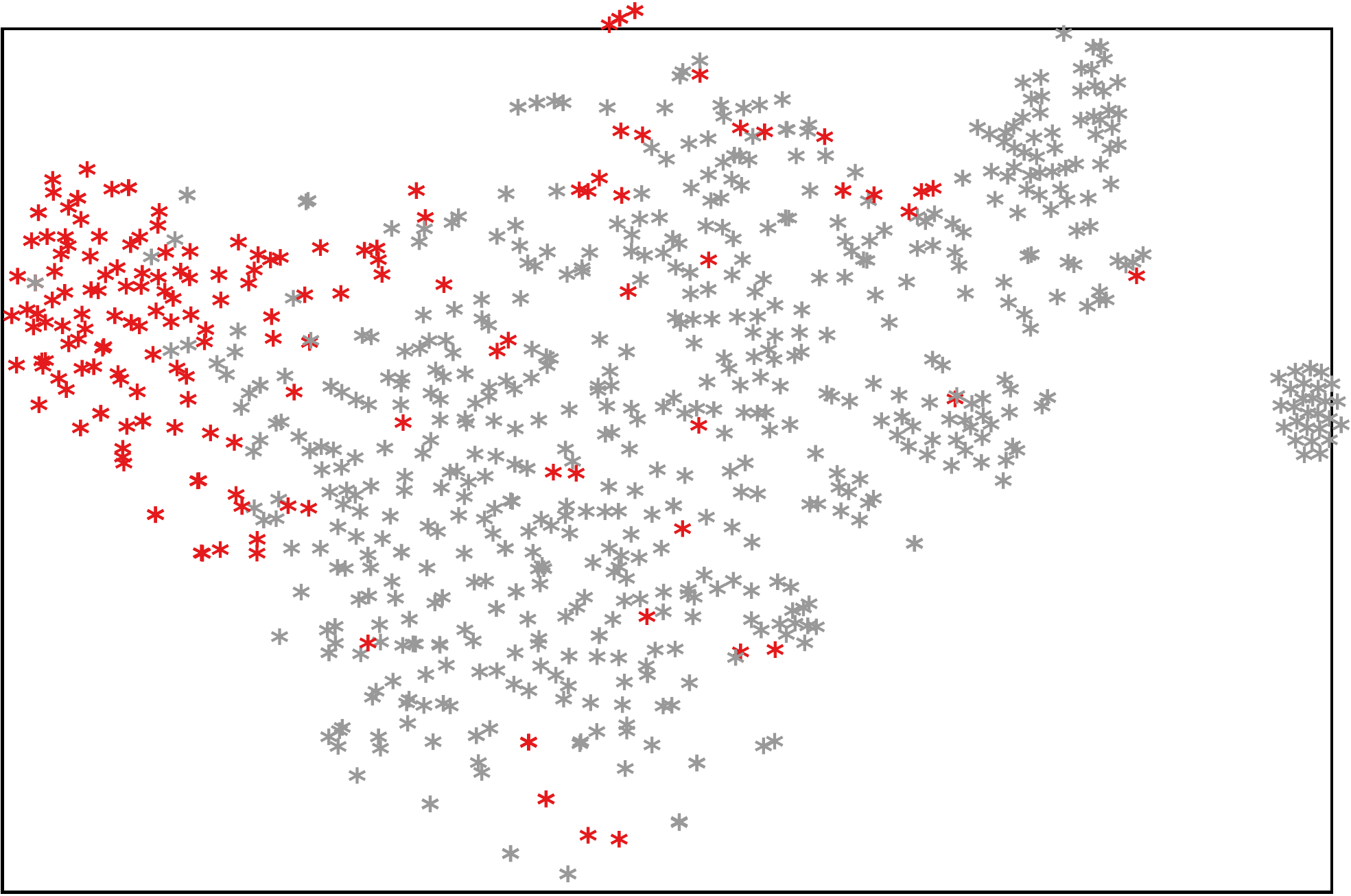}
	\caption{t-SNE visualizations of hidden states for the best-performing query 164 ``\textit{lindsey vonn sidelined}'' from the basic BiCNN (left), QAtt (middle), and PAtt (right). Red (grey) dots represent relevant (non-relevant) posts.}
	\label{vis-158-qatt}
\end{figure*}

\smallskip \noindent \textbf{Per-Query Analysis.} 
In Figure~\ref{per-topic-2013}, we show per-query AP differences between the PAtt model and the QL baseline on the TREC 2013 dataset.
As we can see, PAtt improves on most of the queries.
For the best-performing query 164 ``\textit{lindsey vonn sidelined}'', we project the hidden states $\textbf{o}$ into a low-dimensional space using t-SNE \cite{maaten2008visualizing}, shown in Figure~\ref{vis-158-qatt}.
We observe that with the basic BiCNN model (left), relevant posts are scattered.
With the addition of an attention mechanism (either QAtt in the middle or PAtt on the right), most of the relevant posts are clustered together and separated from the non-relevant posts.
With PAtt, there appears to be tighter clustering and better separation of the relevant posts from the non-relevant posts, giving rise to a better ranking.
We confirmed similar behavior in many queries, which illustrates the ability of our position-aware attention encoder to learn better query-biased representations compared to the other two models.

\begin{table}[t]
	\center
	\small
	\begin{tabular}{p{10em} r  }        \toprule
		\textbf{Match} & \textbf{Count}    \\ \toprule
		\textbf{Oscars}   & 28     \\ 
		\textbf{snub} & 20 \\ 
		\textbf{Affleck}  & 25    \\ 
		\textbf{Oscars snub}   & 18 \\ 
		\textbf{snub Affleck}   & 15 \\ 
		\textbf{Oscars Affleck}   & 23 \\ 
		\textbf{Oscars snub Affleck}   & 13 \\ 
		\bottomrule
	\end{tabular}
	\caption{Matching patterns for the worst-performing query 127 ``\textit{Oscars snub Affleck}''.}
	\label{error-analysis}
\end{table}

For the worst-performing query 125 ``\textit{Oscars snub Affleck}'', the PAtt model lost 0.47 in AP and 0.11 in P30.
To diagnose what went wrong, we sampled the top 30 posts ranked by the PAtt model and counted the number of posts that contain different combinations of the query terms in Table~\ref{error-analysis}.
The PAtt model indeed captures matching patterns, mostly on \textit{Oscars} and \textit{Affleck}. 
However, from the relevance judgments we see that \textit{snub} is the dominant term in most relevant posts, while \textit{Oscars} is often expressed implicitly.
For example, QL assigns more weight to the term \textit{snub} in the relevant post ``\textit{argo wins retributions for the snub of ben affleck}'' because of the term's rarity;
in contrast, the position-aware encoder places emphasis on the wrong query terms.

\smallskip \noindent \textbf{Model Performance.} 
Finally, in terms of training and inference speed, we compared the PAtt model with MP-HCNN on a machine with a GeForce GTX 1080 GPU (batch size:\ 300).
In addition to being more effective (as the above results show), PAtt is also approximately 4$\times$ faster.

\section{Conclusions}

In this paper, we proposed two novel attention-based convolutional encoders to
incorporate query-aware and position-aware information with minimal additional model complexity.
Results show that our model is simpler, faster, {\it and} more effective than previous neural models for searching social media posts.

\section*{Acknowledgments}

This research was supported in part by the Natural Sciences and Engineering Research Council (NSERC) of Canada.

\end{document}